\documentclass[sigconf]{acmart}
\usepackage{multirow}
\usepackage{subcaption}
\usepackage{svg}
\usepackage{float}
\usepackage{graphicx}
\usepackage{enumitem}
\usepackage{placeins}
\usepackage{array}

\AtBeginDocument{%
  }

\copyrightyear{2026}
\acmYear{2026}
\setcopyright{cc}
\setcctype{by-nc-nd}
\acmConference[KDD '26]{Proceedings of the 32nd ACM SIGKDD Conference on Knowledge Discovery and Data Mining V.2}{August 09--13, 2026}{Jeju Island, Republic of Korea}
\acmBooktitle{Proceedings of the 32nd ACM SIGKDD Conference on Knowledge Discovery and Data Mining V.2 (KDD '26), August 09--13, 2026, Jeju Island, Republic of Korea}
\acmDOI{10.1145/3770855.3817978}
\acmISBN{979-8-4007-2259-2/2026/08}

\begin{document}

\author{Xin Wang}
\email{xin.wang.9@stonybrook.edu}
\affiliation{%
  \institution{Stony Brook University}
  \city{Stony Brook}
  \state{New York}
  \country{USA}
}

\author{Yunshi Wen}
\email{yunshi-wen@outlook.com}
\affiliation{%
  \institution{Rensselaer Polytechnic Institute}
  \city{Troy}
  \state{New York}
  \country{USA}
}

\author{Yanan He}
\email{yanan.he@yale.edu}
\affiliation{%
  \institution{Yale University}
  \city{New Haven}
  \state{Connecticut}
  \country{USA}
}

\author{Haotian Xu}
\email{haotian.xu@stonybrook.edu}
\affiliation{%
  \institution{Stony Brook University}
  \city{Stony Brook}
  \state{New York}
  \country{USA}
}

\author{Youlan Zhao}
\email{youlan.zhao@stonybrook.edu}
\affiliation{%
  \institution{Stony Brook University}
  \city{Stony Brook}
  \state{New York}
  \country{USA}
}

\author{Michel Ferreira Cardia Haddad}
\email{m.haddad@qmul.ac.uk}
\affiliation{%
  \institution{Queen Mary University of London}
  \city{London}
  \country{United Kingdom}
}

\author{Tengfei Ma}
\email{tengfei.ma@stonybrook.edu}
\affiliation{%
  \institution{Stony Brook University}
  \city{Stony Brook}
  \state{New York}
  \country{USA}
}

\title[CAAD: Causality-Aware Multivariate Time Series Anomaly
Detection]{CAAD: Causality-Aware Multivariate Time Series Anomaly Detection via Multi-Scale Alignment and Structural Causal Consistency}

\begin{abstract}
  The operational integrity of complex industrial systems relies on precise anomaly detection and diagnosis. The vast majority of existing methods narrowly focus on capturing temporal similarities of representations, often overlooking the disruption of internal causal relationships, which characterizes system failures and latent anomalies. In this paper, we propose a novel framework (CAAD) that reframes anomaly detection as the continuous verification of Granger causality consistency through exogenous variables.
Specifically, the CAAD framework models exogenous time-series variables as residuals, identifying anomalies as significant deviations caused by external interventions. The proposed framework leverages multi-scale alignment to internalize system dynamics and utilizes a gradient-based matrix to monitor internal causal relationship breakdowns. By quantifying causal deviations of both dynamic evolution and relational topology, the CAAD is able to capture subtle causal shifts to achieve precise anomaly detection. Extensive experiments on real-world industrial datasets demonstrate that the CAAD achieves high-precision anomaly detection, outperforming most state-of-the-art baselines.

\end{abstract}

\ccsdesc[500]{Computing methodologies~Machine learning}

\keywords{Time Series; Anomaly Detection; Causality; Multi-Scale Alignment}

\renewcommand{\shortauthors}{Xin Wang et al.}
\maketitle
\section{Introduction}

The reliable operation of complex industrial systems, such as smart grids, water treatment facilities, and large-scale cloud infrastructures, critically depends on accurate multivariate time series anomaly detection. Failures in these systems can propagate rapidly and lead to severe safety, economic, or societal consequences, making early and reliable anomaly identification indispensable \cite{tuli2022tranad,10391270}. In recent years, deep learning methods have been increasingly explored for this task, with approaches largely centered around reconstruction, prediction, or representation association objectives \cite{zamanzadeh2024deep,jin2024survey,boniol2025vus,feng2024sensitivehue,ghorbani2024pate,wang2024cutaddpaste,yu2024pre,wang2024revisiting,nam2024breaking,zhao2024weakly}. Despite their empirical success, most existing methods fundamentally rely on detecting \emph{magnitude-level deviations} or \emph{temporal similarity shifts} in observed signals.

However, real-world system failures are often characterized not by abrupt numerical spikes, but by breakdowns in the underlying \emph{causal logic} that governs system behavior. Since the violation of internal causal relationships often emerges before they manifest large-amplitude signal changes, purely magnitude-driven detectors struggle with \emph{stealthy anomalies} where causal dependencies drift within nominal ranges.

Recognizing its importance, recent studies have attempted to model inter-variable dependencies using graph neural networks or attention mechanisms\cite{zheng2023correlation,chen2021learning,assaad2023root,li2022causal,zhang2021cloudrca,ikram2022root,wang2023interdependent,orchard2025root}. While these methods improve correlation modeling and facilitate dependency-aware representations, they largely remain \emph{correlation-centric}. The learned graphs or attention weights primarily encode co-movement patterns, and anomaly detection is still driven by reconstruction or prediction errors. As a result, these approaches are hypersensitive to noise but insensitive to violations of causal logic.

Parallel to this stream of research, Granger causality has been explored in time series analysis and root cause diagnosis \cite{granger1969investigating,shojaie2022granger,tank2021neural,han2025root,liu2025gcad}. Recent causality-aware models attempt to integrate causal discovery into anomaly detection pipelines \cite{kim2025causality,cho2025structured}. Nonetheless, existing methods typically (i) estimate \emph{static or slowly varying causal graphs}, (ii) treat causality as an auxiliary regularization rather than a primary detection signal, or (iii) focus on post-hoc root cause analysis rather than online anomaly detection. Consequently, they lack a mechanism to continuously verify whether causal relationships remain consistent across time and across multiple temporal resolutions during inference.

In the present work, we argue that anomaly detection in complex systems should be reframed as a problem of \emph{continuous causal consistency verification}. We introduce the \textbf{CAAD} (\textbf{C}ausality-\textbf{A}ware \textbf{A}nomaly \textbf{D}etection), a novel framework that explicitly monitors whether the internal causal structure of a system remains intact as new observations arrive. The CAAD is motivated by two key observations. Firstly, system anomalies are often preceded by causal breakdowns rather than large numerical deviations. Secondly, the underlying dependencies remain invariant across different temporal resolutions, although the numerical patterns exhibit a heterogeneous dynamic.


To operationalize these insights, the CAAD introduces three core innovations:
\begin{enumerate}[nosep, leftmargin=*]
    \item \textbf{Multi-scale Temporal Alignment}: We design a hierarchical temporal modeling architecture that jointly captures fine-grained dynamics and coarse-grained trends. An asymmetric self-supervised alignment objective enforces semantic consistency across scales, enabling robust disentanglement of transient noise from structural behavior.
    
    \item \textbf{Continuous Causal Verification}: Instead of relying on static causal graphs, the CAAD leverages gradient-based Granger signals to continuously assess causal influence among variables during inference. This enables real-time detection of causal relationship breakdowns, even in the absence of large-amplitude signal changes.
    
    \item \textbf{Dual-perspective Anomaly Scoring}: The CAAD combines a dynamic causal score (derived from prediction residuals) with a relational causal score to separately quantify causal dependency from temporal and structural perspectives. This asymmetric fusion allows the causal score to act as a rectifier—amplifying sensitivity to stealthy anomalies, while preserving robustness to local noise and benign fluctuations.
\end{enumerate}

We evaluate the CAAD framework on multiple real-world industrial benchmarks, including SWaT \cite{7469060} and PSM \cite{abdulaal2021practical}. Experimental results demonstrate that the CAAD consistently outperforms state-of-the-art baselines, particularly in scenarios characterized by strong physical constraints and subtle failure modes. On the SWaT dataset, the CAAD achieves an $F_1$ score of $0.95$, substantially exceeding existing correlation and reconstruction-based methods. These results underscore the importance of integrating multi-scale causal awareness as a first-class signal in anomaly detection for complex systems.

\section{Related Work}

\textbf{Multivariate Time Series Anomaly Detection.}
Deep learning has emerged as a predominant paradigm for multivariate time series anomaly detection~\cite{schmidl2022anomaly}, primarily operating under reconstruction-based~\cite{su2019robust, audibert2020usad}, prediction-based~\cite{tuli2022tranad}, or association-based mechanisms~\cite{xu2022anomaly}. To capture the complex interactions among sensors, recent approaches incorporate graph neural networks (GNNs) or attention mechanisms, including GDN~\cite{deng2021graph} and MTAD-GAT~\cite{zhao2020multivariate}, to explicitly model the inter-variable dependency structures and relational correlations. Furthermore, recent advances in general time series analysis have demonstrated that multi-scale modeling is highly effective for disentangling complex temporal variations, such as mixing periodicities and trends, to learn robust representations across diverse time-series tasks, notably TimesNet~\cite{wu2023timesnet} and TimeMixer~\cite{wang2024timemixer, timemixer++}. However, despite these advances in representation learning and dependency modeling, existing methods predominantly rely on point-wise numerical deviations or representation discrepancies as the anomaly criterion, often overlooking the explicit verification of the system's underlying physical or causal consistency. 

\textbf{Causal Discovery and Structure Learning.}
Granger causality has been widely adopted as a practical framework for identifying directed temporal dependencies in monitoring systems, serving as a distinct alternative to broader interventional structural causal models~\cite{shojaie2022granger}. To overcome the scalability issues of traditional combinatorial algorithms, recent research has pivoted toward differentiable approaches, leveraging differentiable optimization and neural parameterizations to efficiently extract interpretable causal graphs from non-linear dynamics~\cite{pamfil2020dynotears, tank2021neural}. Recently, these gradient-based Granger signals have been extended to multivariate anomaly detection and root cause analysis, such as AERCA~\cite{han2025root} and GCAD~\cite{liu2025gcad}. Concurrently, other approaches focus on explicitly modeling structured temporal causality for interpretability~\cite{cho2025structured} or employing causality-aware contrastive learning to enhance robustness~\cite{kim2025causality}. However, these methods tend to either (i) rely on static graph estimation or post-hoc diagnosis, or (ii) utilize causality merely as an auxiliary regularization, often lacking the capability to explicitly verify instantaneous causal consistency across multiple scales during online inference. In contrast, in the present work, we adopt a continuous verification perspective, using gradient-based Granger signals to test structural consistency online across multiple temporal scales.

\begin{figure*}[t]
    \centering
    \includegraphics[
        width=0.95\textwidth,
        trim=30 120 30 0,
        clip
    ]{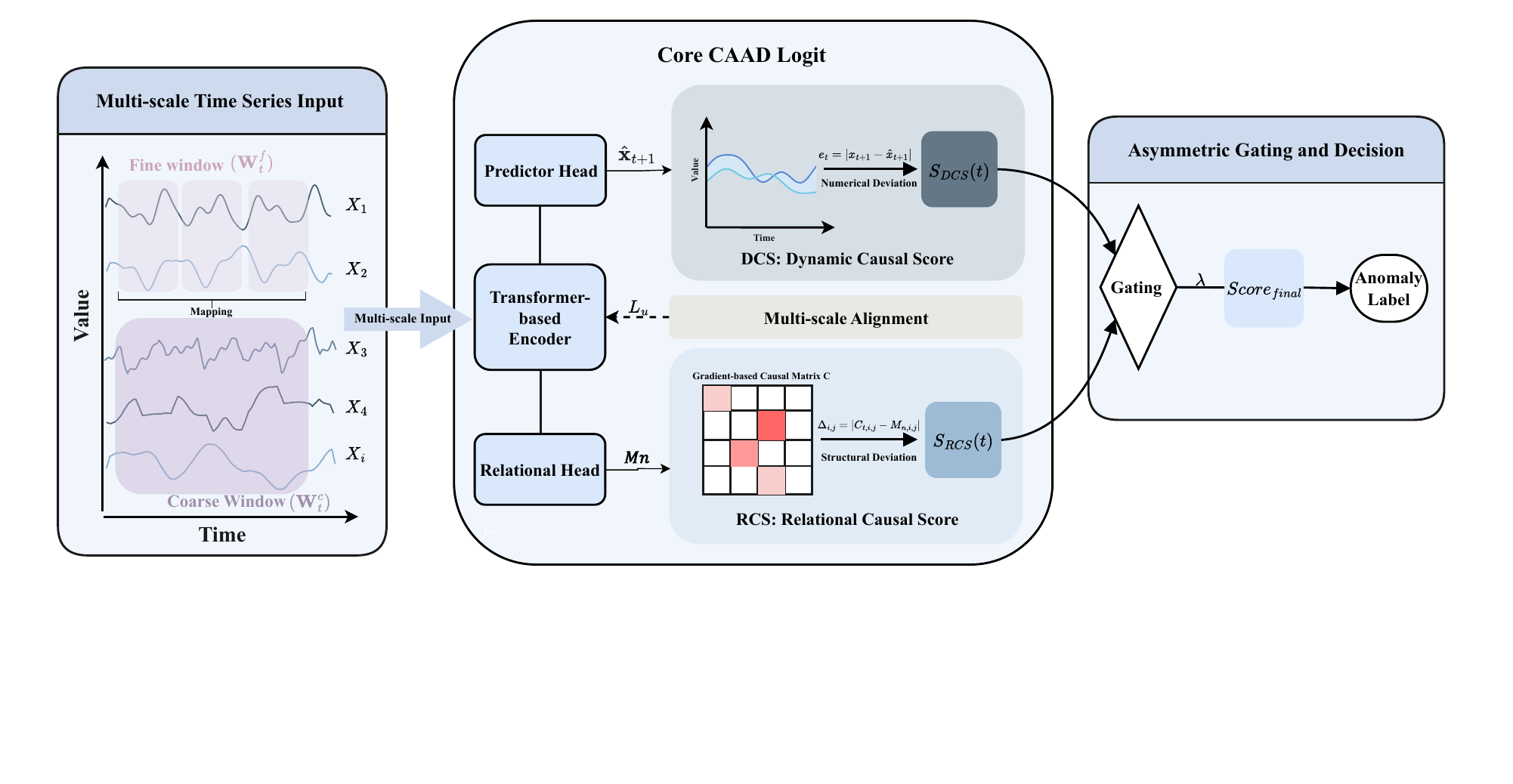}
    \caption{The overall architecture of the CAAD framework, illustrating how multiscale alignment synchronizes causal logic for dual-view deviation quantification ($S_{DCS}$ and $S_{RCS}$).}
    \label{fig:pipeline}
\end{figure*}

\section{Methodology}

\subsection{Problem Statement}
In this paper, we focus on quantifying the causal deviation from the learned normal patterns to detect anomalies present in the multivariate time series. 

Our training data consists of $d$-dimensional multivariate time series, denoted $\mathbf{X} = \{\mathbf{x}_1, \mathbf{x}_2, \dots, \mathbf{x}_T\}$, where each time step $\mathbf{x}_t \in \mathbb{R}^d$ represents the observation of $d$ variates recorded at time $t$. We assume that the training data only consists of normal data, aggregating raw data into fine-grained $\mathbf{X}_f$ and coarse-grained $\mathbf{X}_c$. The input of our model are multi-scale sliding windows $(\mathbf{W}_t^f, \mathbf{W}_t^c)$ ($\mathbf{W}_t = [\mathbf{x}_{t-w+1}, \dots, \mathbf{x}_t]$). The output of our model is a set of anomaly labels $a_{t+1} \in \{0, 1\}$ to indicate whether time step $t+1$ is anomalous.



\subsection{Overview}
Complex dynamical systems, ranging from ecosystems to financial markets to climate, can have tipping points at which a sudden shift to a contrasting dynamical regime may occur \cite{scheffer2009early}. In real-world cyber-physical systems (CPS), causal relationships governed by physical laws often exist between sensors. For example, in the SWaT (secure water treatment) dataset \cite{7469060}, the state of a water valve directly dictates the water level, which subsequently shifts the pH balance. The numerical stability in CPS is a reflection of the underlying causal relational consistency. 

However, conventional methods are mainly limited to magnitude-level deviations while constrained by a single-scale perspective, resulting in a performance bottleneck. 

Therefore, the CAAD aims to capture the normal causal logic through multi-scale time series data, detecting the causal shifts from both dynamic evolution and relational topology perspectives, and identifying the anomalies based on the asymmetric fused anomaly score. The proposed framework consists of the following three primary components:

\begin{enumerate}[nosep, leftmargin=*]
  \item \textbf{Multi-scale Temporal Modeling:} Captures underlying physical causal logic through the alignment of multiple temporal resolutions and computed residuals as proxies for exogenous variables\cite{han2025root}.
  \item \textbf{Dual-view Deviation Quantification:} Utilizes prediction residuals and gradient-based Granger causality matrix, quantifying the causal shifts from dynamic and relational deviation.
  \item \textbf{Asymmetric Gating and Decision:} Employs asymmetric fusion to calculate final anomaly scores to identify anomalies.
\end{enumerate}

\subsection{Multi-scale Temporal Modeling}
While numerical observations in multivariate time series often exhibit multi-scale volatility, the underlying governing laws induced by causal relationships are expected to remain scale-invariant under regular operating conditions. However, capturing these relationships at a single resolution presents a trade-off: fine-grained representations are sensitive to instantaneous causal disruptions (e.g., a sudden valve failure) but are prone to high-frequency stochastic noise. Conversely, coarse-grained representations capture stable causal trends and systemic dependencies but may dilute the localized "intervention" signals that characterize early-stage anomalies. To address this difficulty, we propose a multi-scale temporal modeling approach designed to verify causal consistency across varying temporal abstractions, thereby balancing detection sensitivity with structural stability.

To leverage these complementary strengths and extract a robust causal signature, we employ temporal aggregation to construct a hierarchical input space. Specifically, we define fine-grained windows to preserve high-fidelity local dynamics, and coarse-grained windows generated via a configurable aggregation operator (e.g., mean or max pooling) to represent long-range dependencies. This hierarchical structure allows the model to learn a "causal consensus", where the representation at one scale provides a context-aware prior for the other, enhancing the ability of the model to distinguish between random noise and genuine structural breakdowns \cite{muttenthaler2025aligning}.

Due to the fact that temporal resolution differs across scales, we apply a mapping function $\mathcal{M}$ that properly maps each fine-grained window to its coarse-grained window. This ensures temporally synchronized alignment and a rigorous foundation for subsequent cross-scale learning.




\textbf{Asymmetric Alignment Mechanism.}
Our framework utilizes two transformer encoders to separately extract temporal features from both fine and coarse windows, deriving window-level latent embeddings $\mathbf{h}_t^{(f)}$ and $\mathbf{h}_t^{(c)}$. In prediction, for each scale, the predictor $g(\cdot)$ first generates base prediction vectors $\tilde{y} \in \mathbb{R}^{1 \times d}$ for all variables via a two-layer multilayer perceptron (MLP). To explicitly model inter-variable dependencies, we introduce a learnable relation matrix $A \in \mathbb{R}^{d \times d}$ that performs a linear transformation on these vectors, yielding the final prediction:
\begin{equation}
    \hat{y} = \tilde{y}A^\top
\end{equation}
Since we assume that underlying structural relationships within industrial systems remain invariant in the normal training dataset, the learnable relation matrix $A$ is shared across multiple scales, which maintains consistency and incorporates structural prior across different scales.


Strict numerical alignment tends to eliminate the meaningful differences between scales, causing individual branches to lose their unique informative value. Therefore, rather than forcing direct numerical alignment, we adopt an \textit{asymmetric self-supervised alignment architecture} inspired by the algorithm BYOL (bootstrap your own latent) \cite{grill2020bootstrap}, on two scales: Hidden representations $\mathbf{h}_t^{(f)}$ and $\mathbf{h}_t^{(c)}$. In practice, because coarse-grained representations are more robust against local noise after aggregation, the fine-grained ``student'' branch employs an MLP predictor $q$ to map its representations to the coarse-grained ``teacher'' space and generates
\begin{equation}
    p_{f} = q(\mathbf{h}_t^{(f)})
\end{equation}

To prevent representation collapses, the teacher branch remains stationary via a stop-gradient mechanism. The student’s predicted representation $p_{f}$ and the teacher’s target representation $z_{c}$ are projected onto a unit hypersphere to eliminate scale-specific magnitude variances and focus on directional alignment:

\begin{equation}
    L_u = \left\| \frac{p_{f}}{\|p_{f}\|_2} - \text{stop\_grad} \left( \frac{z_{c}}{\|z_{c}\|_2} \right) \right\|_2^2
\end{equation}

The overall training loss is $\mathcal{L}$, the sum of multi-scale forecasting errors and \textit{the alignment loss $L_u$}, where $L_u = \text{MSE}(p_{f}, z_{c})$. The model ultimately outputs predicted values for all variables at the next time step, $\mathbf{x}_{t+1} \in \mathbb{R}^d$ on the coarse scale. 
\begin{equation}
    \mathcal{L} = \text{MSE}_{fine} + \text{MSE}_{coarse} + \alpha \cdot L_u
\end{equation}

By decoupling structural dynamics from transient noise, this hierarchical alignment provides a reliable foundation for causal discovery. While causal dependencies are often stable in physical systems, random noise and sensor errors tend to be transient and fluctuate unpredictably across different scales. By enforcing consistency between the fine and coarse representations, our model effectively filters out such spurious variations, isolating the stable system dynamics while facilitating more accurate anomaly detection.

\subsection{Dual-view Deviation Quantification}
By establishing a consistent cross-scale causal representation through the alignment module, the challenge turns to how to accurately extract anomaly signals and identify anomalies. Following the Granger causal framework \cite{han2025root}, system observations depend on past dynamics plus an exogenous term; prediction residuals serve as proxies for this term, remaining small under normal causal dynamics and amplifying upon anomalous disruptions. Thus, we capture causal logic shifts and investigate the final anomaly score along two dimensions: \textit{dynamic causal score (DCS)} and \textit{relational causal score (RCS)}.


\textbf{Dynamic Causal Score.}
The prediction errors produced in the module of multi-scale temporal modeling directly quantify the system's temporal causal deviation from established nominal patterns. By adopting residuals as the primary metric, we evolve the task of anomaly detection from simply locating outliers to identifying logical inconsistencies. 

In the test dataset, given the actual observation in the next time step $\mathbf{x}_{t+1} \in \mathbb{R}^d$ and the corresponding prediction $\hat{\mathbf{x}}_{t+1}$ generated by our pre-trained framework $g(\cdot)$, we define the residual vector $e_t$ as the absolute discrepancy between the actual observation and the model predicted value at time $t$:
\begin{equation}
    e_t = |\mathbf{x}_{t+1} - \hat{\mathbf{x}}_{t+1}|
\end{equation}

Given the scale heterogeneity among sensors, we normalize residuals using robust statistics. For each variable $i$, robust estimators, median (${median}(r_i)$) and median absolute deviation (MAD) (${MAD}(r_i)$) are calculated on the normal training data for standardization. By scaling the residuals, we mitigate the effect of noise while ensuring that residuals are standardized on a comparable scale across all variables with reduced sensitivity to extreme values:
\begin{equation}
    \tilde{r}_{i,t}
    =
    \frac{
    r_{i,t} - \operatorname{median}(r_i)
    }{
    \operatorname{MAD}(r_i) + \epsilon
    }
\end{equation}

$\tilde r_{i,t}$ represents the temporal deviation of the variable $i$ from its normal reference at time $t$. Lastly, since simple averaging dilutes local anomaly signals and the maximum is susceptible to noise, we implement a top-$K$ strategy $K = \max(1, \lceil d \times 0.1 \rceil)$, determined solely by dimensionality, to derive the final DCS, with robust performance across a range of $K$ values:

\begin{equation}
    S_{\mathrm{DCS}}(t) = \frac{1}{k} \sum_{i \in \mathrm{Top}\text{-}k(\tilde{r}_{t})} \tilde{r}_{i,t}
    \label{eq:dcs}
\end{equation}


\textbf{Granger Causal Discovery and Refinement.}
A prerequisite to deriving the relational causal score (RCS) is the extraction of the instantaneous and refined causal relationships matrix.
Our architecture employs \textit{gradient-based Granger causality} (GC), using the gradients to quantify the contributions of the input variables to the prediction while identifying the underlying causal structure.

For an input window $\mathbf{W}_t = [\mathbf{x}_{t-w+1}, \dots, \mathbf{x}_t]$, we generate predictions using the pre-trained mapping function $g(\cdot)$ that characterizes normal system patterns. To quantify the causal influence between variables, we compute the Jacobian matrix $J_{ij\tau}$. For each $j$-th predicted variable $\hat{x}_j$ with respect to the $i$-th input variable $x_i$ at time lag $\tau$ is captured as:
\begin{equation}
    J_{ij\tau} = \frac{\partial \hat{x}_j}{\partial x_{i,\tau}}
\end{equation}

Here, $J_{ij\tau}$ represents the sensitivity of the $i$-th sensor reading to a change in the $j$-th sensor at lag $\tau$. To ensure that discovered causal links are robust and not based on instantaneous noise, we aggregate these gradient intensities over a temporal window of size $T$. The resulting window-level Granger causality matrix $d \times d$ characterizes the current causal structure within that window $G \in \mathbb{R}^{d \times d}$, defined as:
\begin{equation}
    G_{ij} = \sum_{\tau=1}^{T} \left| \frac{\partial \hat{x}_{j}}{\partial x_{i,\tau}} \right|
    \label{eq:gc}
\end{equation}

While the raw causal graph captures initial gradient sensitivities, it often contains weak or redundant edges that lack physical interpretability. To refine these interactions into a robust structure, we apply two primary refinement strategies, with an optional sparsification step available for additional control.

\begin{enumerate}[nosep, leftmargin=*]
    \item \textbf{Self-correlation Masking}: In time series, a gradient contribution of a variable to its own prediction typically dominates the graph due to strong self-correlation. To ensure the model focuses on cross-variable dependencies, we mask the diagonal elements: $M_{n, ii} = 0$
    \item \textbf{Causal Directional Pruning}: Given asymmetrical causality in real-world settings, we apply directional pruning to resolve bidirectional edges ($A \leftrightarrow B$). By comparing the relative magnitudes of $G_{ij}$ and $G_{ji}$, we retain only the dominant direction. This process restores the unidirectional information flow in real industrial systems.
    \item \textbf{Threshold-based Sparsification}: As an optional refinement, we provide threshold-based sparsification to filter spurious correlations: $M_{n,ij} = \text{Threshold}(G_{ij})$. In practice, self-correlation masking and directional pruning alone yield a robust causal structure sufficient for detection.
    
\end{enumerate}

\textit{Normal Reference Matrix}: Unlike the static relationships captured in the shared relation matrix $A$, the normal reference graph ($M_n$) is constructed by temporally aggregating the refined GC graphs across all normal training windows. This matrix represents the standard variable relationships of the system, serving as the reference for identifying subsequent structural anomalies.

\textbf{Relational Causal Score.}
By applying the refinement steps above, we generate an instantaneous causal matrix $\mathbf{C}_t \in \mathbb{R}^{d \times d}$ for each window during the inference stage. While $\mathbf{C}_t$ reflects the variable interactions within the current window, $\mathbf{M}_n$ serves as the standard reference benchmark for a healthy system.


The RCS evaluates the causal consistency of the industrial systems by first measuring the element-wise discrepancy between the current causal matrix and the reference of the normal pattern: 
\begin{equation}
    \Delta_{ij} = |C_{t,ij} - M_{n,ij}|
\end{equation}
The final relational causal score $S_{RCS}(t)$ is defined as the Frobenius norm of these deviations, providing a secondary validator of the causal logic violation across all sensors:
\begin{equation}
    S_{\mathrm{RCS}}(t)=\|\Delta_t\|_F
\end{equation}

\subsection{Asymmetric Gating and Decision}
Considering the high complexity of real-world system data, it is barely practical to capture diverse anomaly patterns from a single perspective. Consequently, our model introduces a dual-perspective anomaly scoring mechanism. 


In our design, $S_{DCS}(t)$ serves as the primary indicator of anomalies by tracking prediction residuals, while $S_{RCS}(t)$ functions as a secondary validator of system internal causal logic:
\begin{equation}
    Score_{final} = S_{DCS}(t) + \lambda \cdot \text{ReLU}(S_{RCS}(t) - \tau)
    \label{eq:gating}
\end{equation}
Here, $\tau$ is a prior percentile that defines the boundary of normal patterns. In practice, we find $\tau$ to be remarkably robust; empirical evaluations indicate that the performance of CAAD remains highly stable when the threshold $\tau \in [q_{70}, q_{99}]$. Therefore, we set it as the 90th percentile ($q_{90}$). This implies that only the most extreme 10\% of RCS scores are considered indicators of severe causal structure collapses. The $\lambda$ controls the degree of contribution from the RCS to the final anomaly score. This trigger mechanism is designed for: 
\begin{enumerate}[nosep, leftmargin=*]
\item \textbf{Stealthy Anomalies}: Numerical deviation in $S_{DCS}(t)$ may remain negligible despite a dramatic shift in the causal relationships between variables. In such scenarios, the structural score $S_{RCS}(t)$ provides a compensatory structural evidence for the final decision-making. 
\item \textbf{False Alarms}: Following an anomaly event, the numerical residuals in $S_{DCS}(t)$ may continue to fluctuate above the threshold even though the causal relationship stays consistent. In such scenarios, $S_{RCS}(t)$ acts as a secondary validator that constrains the final anomaly score and effectively filters out these false alarms.

\end{enumerate}

\textit{Root Cause Diagnosis}: Although the main objective of this paper is focused on anomaly detection, our architecture is naturally extendable to include the root cause diagnosis. Unlike traditional models with undirected and static graphs that are unable to attribute and localize the root cause, our model supports directed and real-time gradient-based causal matrix, which can be aggregated into interpretable root‑cause scores. Therefore, we leave the root cause diagnosis for further exploration in the future.


\begin{table*}[t]
\centering
\small
\setlength{\tabcolsep}{3.5pt}
\begin{tabular}{ll|cc|ccccc|c}
\hline
Dataset & Metric & PCA & KNN & iTransformer & PatchTST & TranAD & VAE & TimesNet & \textbf{CAAD} \\
\hline
\multirow{3}{*}{SWaT}
& F1   & 0.261 & 0.217 & 0.039 & 0.066 & 0.323 & \underline{0.324} & 0.082 & \textbf{0.952} \\
& PRC & 0.504 & 0.507 & 0.116 & 0.121 & 0.498 & \underline{0.575} & 0.175 & \textbf{0.974} \\
& ROC& 0.687 & 0.719 & 0.336 & 0.343 & 0.691 & \underline{0.745} & 0.392 & \textbf{0.994} \\
\hline
\multirow{3}{*}{PSM}
& F1   & \underline{0.424} & \textbf{0.436} & 0.109 & 0.111 & 0.405 & 0.011 & 0.117 & 0.362 \\
& PRC & 0.413 & \underline{0.456} & 0.376 & 0.377 & 0.438 & 0.393 & 0.360 & \textbf{0.602} \\
& ROC& 0.576 & \underline{0.667} & 0.582 & 0.581 & 0.574 & 0.586 & 0.558 & \textbf{0.739} \\
\hline
\multirow{3}{*}{CATSv2}
& F1   & 0.086 & 0.060 & 0.073 & 0.133 & 0.139 & 0.079 & \underline{0.262} & \textbf{0.330} \\
& PRC & 0.134 & 0.128 & 0.097 & 0.120 & 0.127 & 0.118 & \underline{0.265} & \textbf{0.321} \\
& ROC& 0.644 & \underline{0.748} & 0.661 & 0.704 & 0.617 & 0.655 & \textbf{0.772} & 0.707 \\
\hline
\multirow{3}{*}{SMD}
& F1   & 0.176 & 0.112 & \underline{0.189} & 0.187 & 0.163 & 0.021 & \textbf{0.191} & 0.170 \\
& PRC & 0.109 & 0.134 & 0.142 & \underline{0.144} & 0.105 & 0.098 & \textbf{0.202} & 0.090 \\
& ROC& 0.655 & 0.716 & \underline{0.758} & 0.755 & 0.669 & 0.641 & \textbf{0.797} & 0.674 \\
\hline
\end{tabular}
\caption{Performance comparison of CAAD and state-of-the-art baselines across four public benchmark datasets, demonstrating that the CAAD consistently outperforms the majority of baselines.}
\label{overall}
\end{table*}

\section{Experiments}

To evaluate the performance of the CAAD framework, we conduct experiments on four multivariate time series anomaly detection datasets: SWaT, PSM, CATSv2, and SMD. These datasets cover industrial control systems, server and cloud monitoring, and controlled synthetic settings.

\subsection{Experimental Setup}
\textbf{Datasets and Baselines.}
We utilize four datasets spanning diverse domains: the Secure Water Treatment (SWaT) \cite{7469060} industrial testbed with physical dependencies, the Pooled Server Metrics (PSM) \cite{abdulaal2021practical} dataset from eBay, the synthetic Controlled Anomalies Time Series (CATSv2) \cite{fleith2023cats} with controlled anomalies, and the Server Machine Dataset (SMD) \cite{su2019omnianomaly} containing multi-week server monitoring data. For SWaT, we select 30 sensors out of 51 total based on information content (variance and dynamic range), excluding low-variance actuator signals. We compare our method with a diverse set of baselines spanning classical and deep learning approaches. Classical methods include PCA \cite{shyu2003novel}, which detects anomalies via reconstruction error in a low-dimensional subspace, and KNN \cite{angiulli2002fast}, which identifies anomalies based on nearest-neighbor distances. Transformer-based forecasting models include iTransformer \cite{ICLR2024_2ea18fdc}, PatchTST \cite{nie2023patchtst}, and TranAD \cite{tuli2022tranad}. We also include VAE \cite{kingma2013auto} as an autoencoder-based baseline and TimesNet \cite{wu2023timesnet}, which captures multi-periodicity in the frequency domain. Methods relying on point-adjusted F1 are excluded from direct comparison, as this protocol systematically inflates scores under sparse anomaly distributions \cite{kim2022towards}.

\textbf{Implementation and Metrics.} 
We split chronologically ordered normal data into 80\% for training and 20\% for validation. Data undergoes $z$-score normalization and windowing into fine/coarse patches, where coarse windows inherit anomaly labels from any constituent fine patch. The model is trained for 50 epochs using AdamW. We evaluate performance via F1-score, AUC-ROC, and AUPRC. Given the highly imbalanced nature of anomaly detection, we prioritize AUPRC for model selection. Thresholds are determined from the training data, with final performance reported on the test set.

\begin{table}[t]
\centering
\small
\setlength{\tabcolsep}{6pt}
\begin{tabular}{l|l|ccc}
\hline
Dataset & Setting & F1 & PRC & AUC \\
\hline
\multirow{2}{*}{SWaT}
& \textbf{CAAD‑MS} 
& \textbf{0.9517} & \textbf{0.9738} & \textbf{0.9936} \\
& CAAD‑SS
& 0.7837& 0.6318& 0.7692\\
\hline
\multirow{2}{*}{PSM}
& \textbf{CAAD‑MS} 
& \textbf{0.3620} & \textbf{0.6020} & \textbf{0.7390}\\
& CAAD‑SS
& 0.2497& 0.5941& 0.7258\\
\hline
\multirow{2}{*}{CATSv2}
& \textbf{CAAD‑MS} 
& \textbf{0.3302} & \textbf{0.3208} & \textbf{0.7071} \\
& CAAD‑SS
& 0.2378& 0.2179& 0.6483\\
\hline
\multirow{2}{*}{SMD}
& \textbf{CAAD‑MS} 
& \textbf{0.1698} & \textbf{0.0892}& \textbf{0.6739} \\
& CAAD‑SS
& 0.0135& 0.0747& 0.6629\\
\hline
\end{tabular}
\caption{Ablation comparison between multi-scale(MS) and single-scale(SS) CAAD, illustrating that MS optimizes causal logic capture, yielding more accurate prediction.}
\label{albation}
\vspace{-2em}
\end{table}

\subsection{Anomaly Detection Performance}

Table \ref{overall} summarizes the performance of the CAAD alongside seven state-of-the-art baselines in four public benchmarks: SWaT, PSM, CATSv2, and SMD. The performance of the baselines is obtained from \cite{qiu2025tab}.

In particular, the CAAD achieves superior results in most cases, most notably on the SWaT dataset. In this high-dimensional setting, statistical methods such as PCA and KNN struggle to capture complex physical interactions, while reconstruction-based models such as VAE and TranAD often fail to exceed the detection threshold. This issue is especially evident under stealthy anomalies, where shifts in causal relationships do not manifest as obvious numerical spikes, leading to high miss rates for standard baselines.

Furthermore, although deep forecasting models such as iTransformer and PatchTST excel at temporal modeling, they are largely magnitude-driven and thus unable to detect shifts in cross-variable logic. The CAAD explicitly addresses this limitation by leveraging dynamic relational causal scores to assess the stability of the underlying physical logic, enabling the detection of deeper, non-numerical anomalies that elude purely temporal or statistical approaches.

Unlike the SWaT dataset with rigid physical dependencies, the SMD consists of server metrics from an internet data center. Due to SMD's relatively sparse causal interactions, the CAAD can be sensitive to misidentifying random noise as structural logic shifts because our model is designed mainly for datasets with obvious structural causal shifts. Thus, in such weakly-coupled scenarios, the performance of TimesNet marginally outperforms the CAAD on the SMD dataset.


\subsection{Ablation Study}

\subsubsection{Effect of Multiscale Feature Synchronization}

We conduct ablation experiments to evaluate the effectiveness of the multi-scale architecture. As shown in Table \ref{albation}, the multi-scale design outperforms the single-scale variant across all benchmarks and metrics. The improvement is most pronounced on datasets with rigid physical dependencies, such as SWaT, where the $F_1$ score reaches 0.95, a 20\% increase over the single-scale version. This is because the multi-scale alignment synchronizes and captures the system's causal logic across different frequencies, leading to a more robust representation compared to the single-scale.


To further investigate the benefits of this design, we analyze the quality of the prediction residuals, which serve as proxies for exogenous variables. 

\begin{figure*}[t]
    \centering
    \begin{subfigure}{0.23\textwidth}
        \centering
        \includegraphics[width=\linewidth]{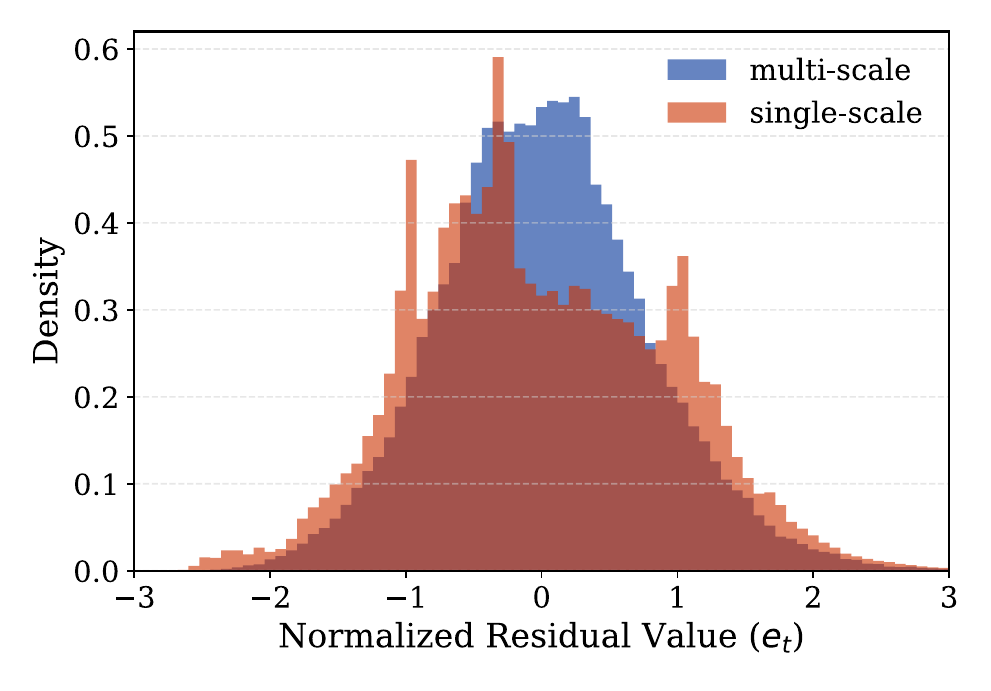}
        \caption{CATSv2}
    \end{subfigure}
    \hfill
    \begin{subfigure}{0.23\textwidth}
        \centering
        \includegraphics[width=\linewidth]{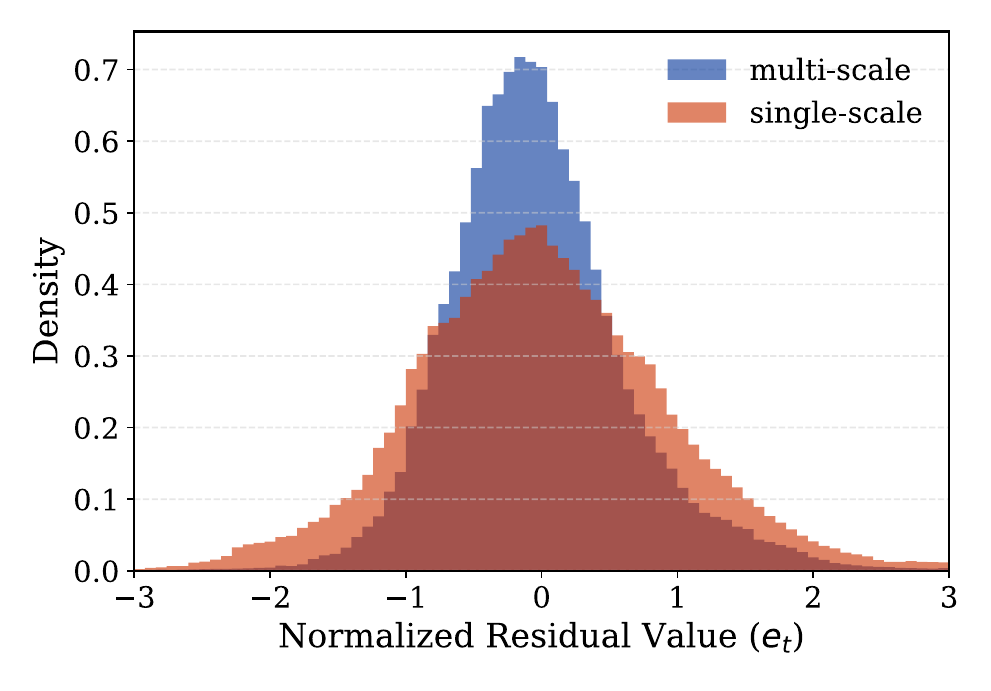}
        \caption{PSM}
    \end{subfigure}
    \hfill
    \begin{subfigure}{0.23\textwidth}
        \centering
        \includegraphics[width=\linewidth]{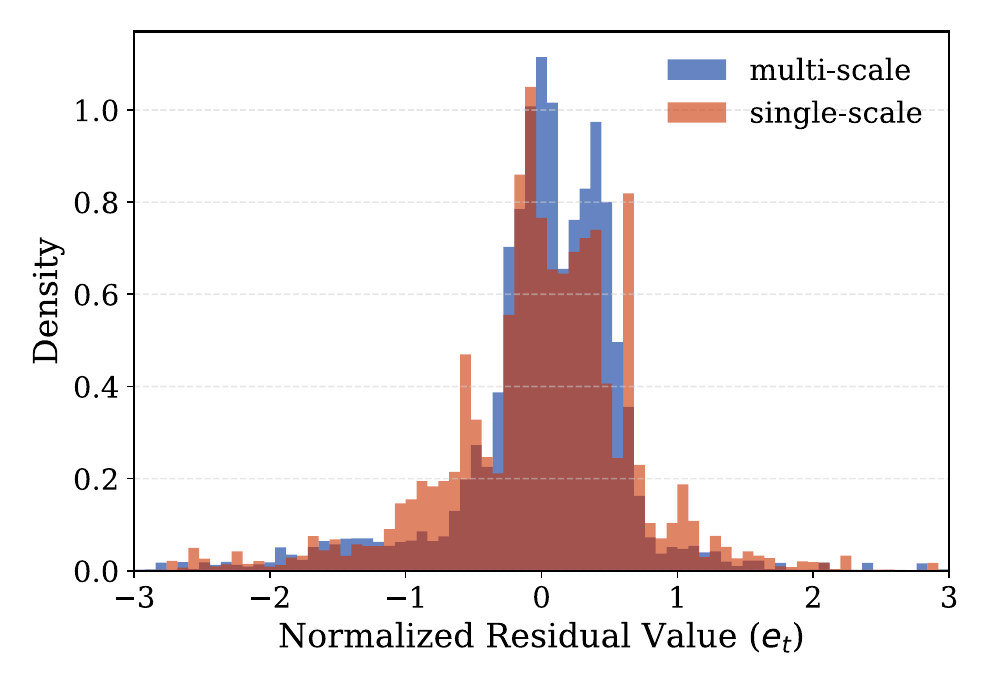}
        \caption{SWaT}
    \end{subfigure}
    \hfill
    \begin{subfigure}{0.23\textwidth}
        \centering
        \includegraphics[width=\linewidth]{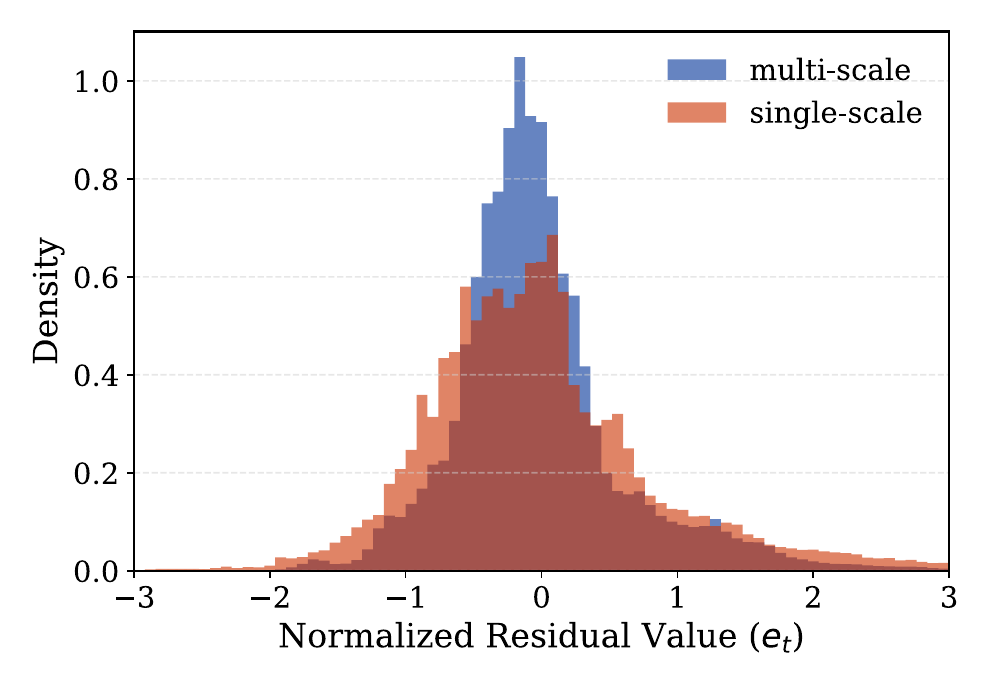}
        \caption{SMD}
    \end{subfigure}

    \caption{ Residual density comparison across all benchmarks: Multi-scale CAAD consistently yields a more leptokurtic distribution (narrower profile with higher kurtosis) than the single-scale version, showing enhanced residual stability.}
    \label{fig:g1_1x4}
\end{figure*}

\begin{figure*}[t]
    \centering
    \begin{subfigure}{0.23\textwidth}
        \centering
        \includegraphics[width=\linewidth]{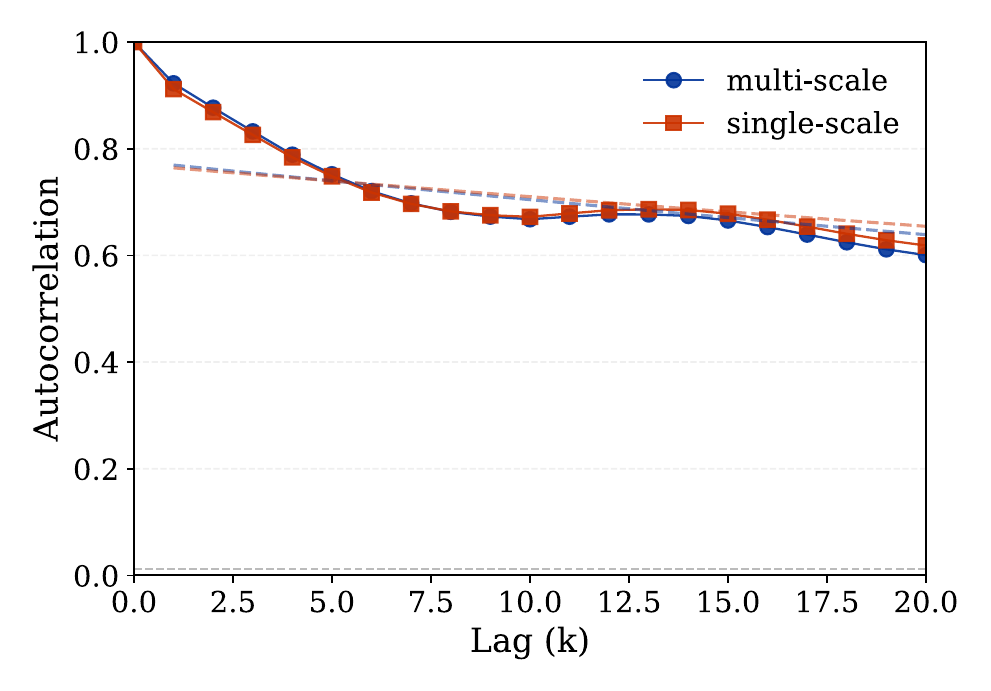}
        \caption{CATSv2}
    \end{subfigure}
    \hfill
    \begin{subfigure}{0.23\textwidth}
        \centering
        \includegraphics[width=\linewidth]{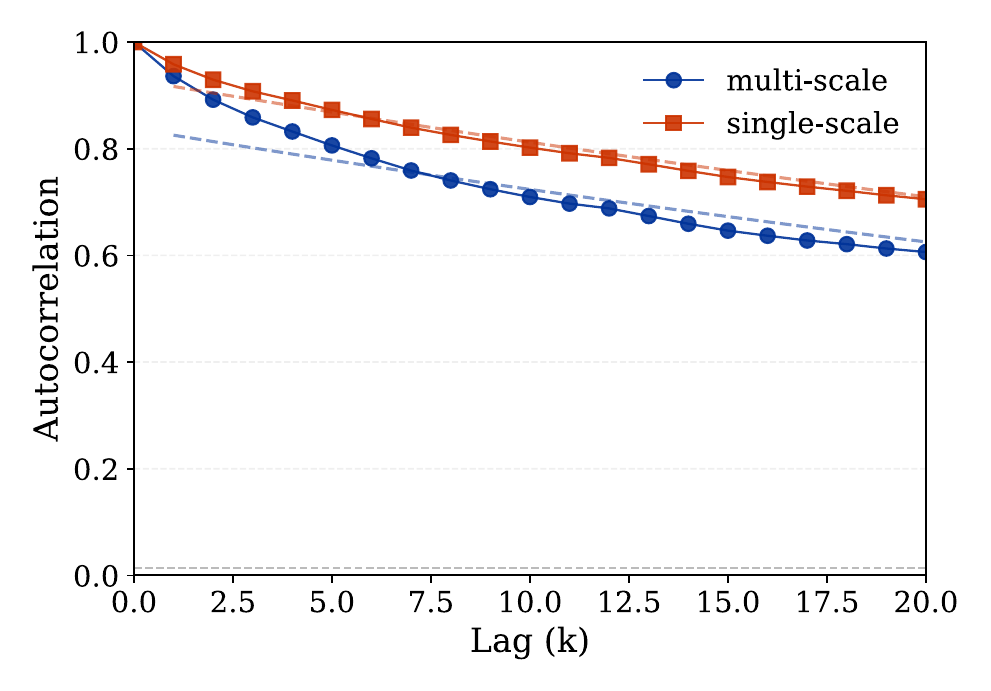}
        \caption{PSM}
    \end{subfigure}
    \hfill
    \begin{subfigure}{0.23\textwidth}
        \centering
        \includegraphics[width=\linewidth]{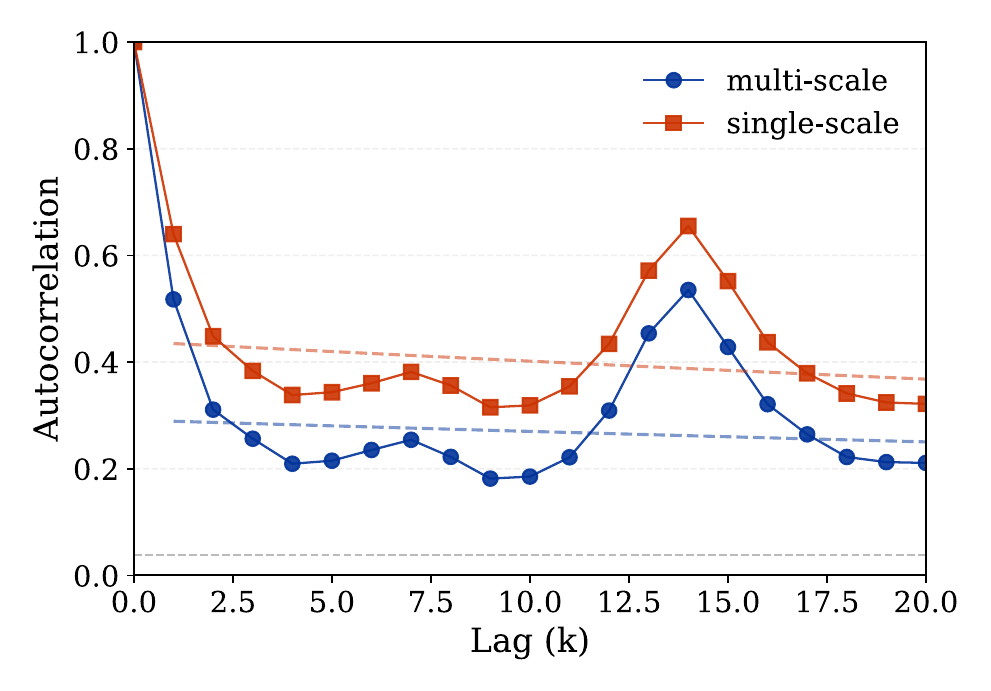}
        \caption{SWaT}
    \end{subfigure}
    \hfill
    \begin{subfigure}{0.23\textwidth}
        \centering
        \includegraphics[width=\linewidth]{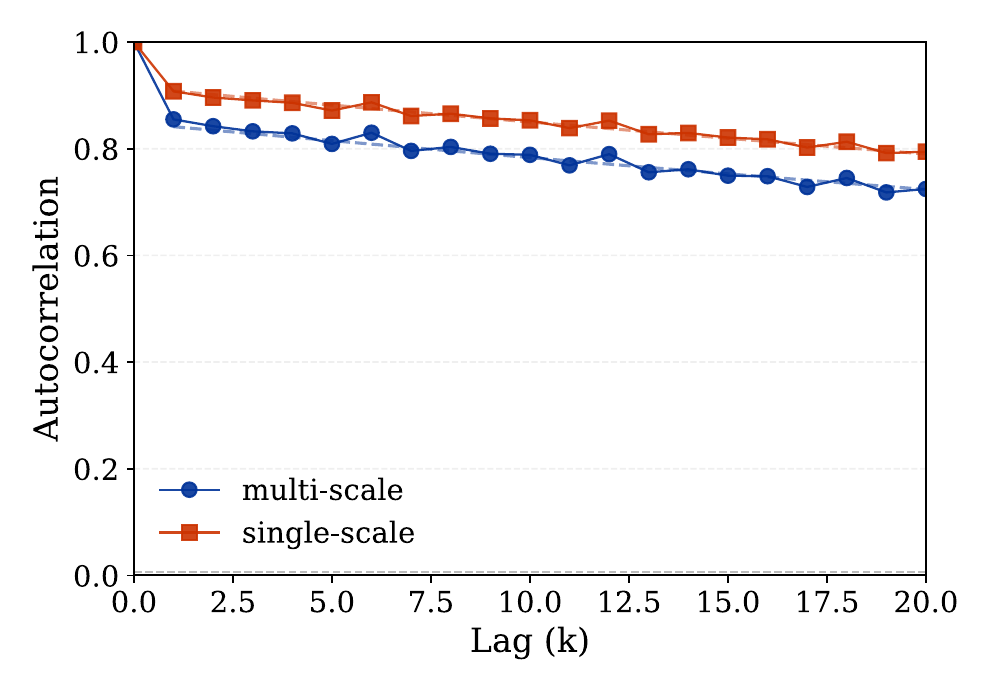}
        \caption{SMD}
    \end{subfigure}

    \caption{Residual Autocorrelation comparison across all benchmarks: Multi-scale CAAD exhibits lower autocorrelation, indicating a more comprehensive capture of cross-scale causal logic.}
    \label{fig:g2_1x4}
\end{figure*}
\textbf{The Distributional Effect.}
The precise identification of causal shifts is based on the quality of these residuals. 
The central limit theorem (CLT) implies that the sum of independent random variables tends toward a Gaussian distribution \cite{hyvarinen2000independent}. In the context of anomaly detection, a more Gaussian residual distribution suggests that the model fails to learn and extract significant causal dependencies during training. Our results show that the multi-scale kurtosis, a standard measure of non-Gaussianity, is significantly higher than that of the single-scale version, even reaching a three-fold difference on the SMD dataset in  Fig. \ref{fig:g1_1x4}

This suggests that single-scale models, limited by a narrow temporal perspective, fail to capture long-term dynamics and leave unmodeled useful signals in the residuals. In contrast, the multi-scale model achieves higher alignment with real-world system dynamics, considering that its residual distribution exhibits a super-Gaussian distribution\cite{xie2025characterizing}. The multi-scale setting also exhibits a smaller standard deviation of residuals during normal windows, which further indicates that the multi-scale design establishes a more robust foundation for the subsequent detection of structural anomalies.

\textbf{The Autocorrelation Effect.}
Beyond static noise reduction, the ability of a model to extract complex temporal dependencies is a critical performance metric. Complex temporal variations are composed of multiple intraperiod and interperiod variations \cite{wu2023timesnet}. Therefore, the remaining periodicity in the residuals is considered as an indicator of how effectively the model decouples causal dynamics. To verify this, we performed a residual whitening test to evaluate temporal autocorrelation. Ideally, residuals should behave as white noise, indicating that all learnable dynamic causal patterns have been successfully extracted.

Figure \ref{fig:g2_1x4} illustrates the auto-correlation function (ACF) curves for our experiments. We observe that in the SWaT dataset, a subtle periodic pattern persists roughly every 14 time steps. Even in high-precision systems such as GNSS positioning, flicker noise is often unavoidable due to the inherent physical nature of the data  \cite{xie2025characterizing}. Despite this, the multi-scale ACF decays significantly faster than the single-scale version, reflecting a more thorough extraction of temporal features.

By quantifying the Ljung-Box $Q$-statistic for SWaT, a formal test for residual whiteness, the number drops by 47.9\% in multi-scale CAAD. These results demonstrate that the model decouples noise from the underlying causal logic and yields more decorrelated residuals by integrating multiscale synchronization.

\begin{figure}[t]
  \centering
  \includegraphics[width=0.7\linewidth]{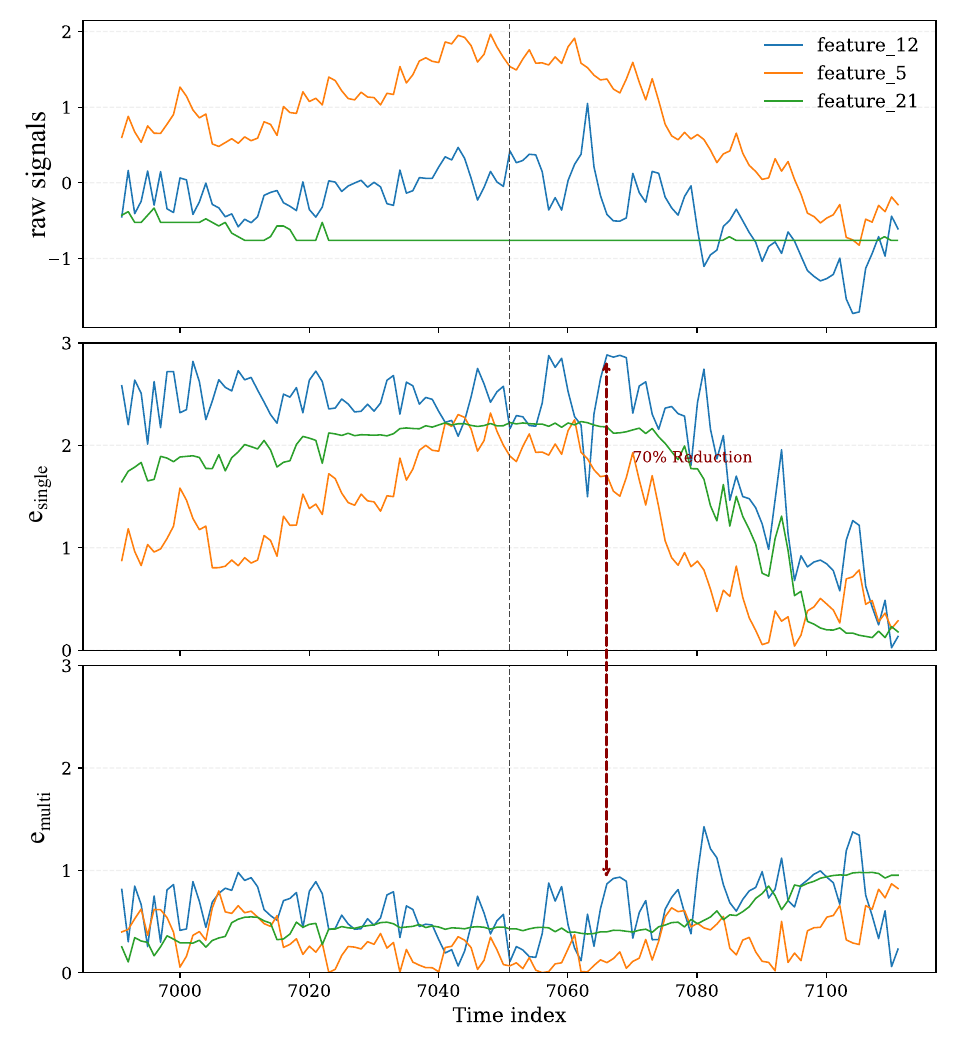}
  \caption{Comparison of raw signals (top) with residuals generated by single-scale (middle) and multi-scale (bottom) CAAD. Multi-scale alignment significantly suppresses spurious fluctuations in normal windows.}
  \label{fig:g3}

\end{figure}

\textbf{The Robustness Effect.}
In industrial deployments, the false alarm rate is one of the reliable measurements of an anomaly detection system. To demonstrate the practical value of the multi-scale design, we present a case study and highlight three variables with the most significant performance gap in Fig. \ref{fig:g3}.

As shown in Fig.\ref{fig:g3}, the single-scale residuals are highly correlated with the input signal, essentially mirroring the raw fluctuations and leading to spurious spikes that inevitably trigger false alarms. In contrast, the multi-scale residuals remain stable and are consistently suppressed below 0.6. At the point of greatest discrepancy, the multi-scale approach achieves a 70\% reduction in residual magnitude compared to the single-scale. This highlights the CAAD's ability to distinguish between genuine anomalies and oscillations, emphasizing its robustness for industrial applications.

\textbf{Hyperparameter Analysis.}
The hyperparameter $\alpha$ is introduced in Section 3.3, controlling the weight of the multi-scale alignment loss. 
To systematically access its impact, we apply a logarithmic interval search on the grid, ranging from ``no alignment'' to ``strong regularization'', which ensures a thorough evaluation of the robustness and stability of CAAD at different intensity levels. 

\begin{figure}[t]
    \centering
    \begin{subfigure}{0.48\linewidth}
        \centering
        \includegraphics[width=\linewidth]{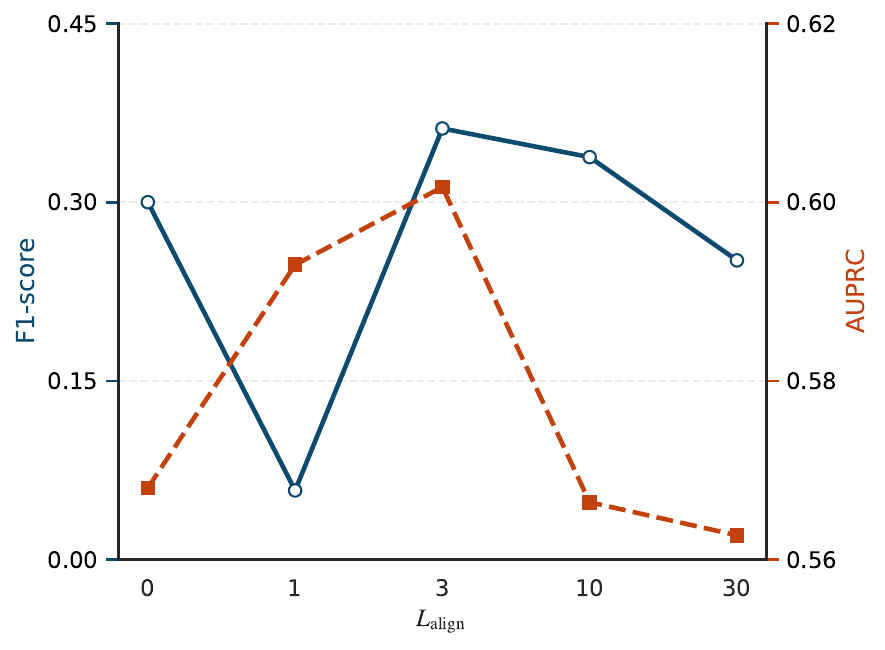}
        \caption{Alignment weight $\alpha$.}
        \label{fig:psm_lu_sensitivity}
    \end{subfigure}
    \hfill
    \begin{subfigure}{0.48\linewidth}
        \centering
        \includegraphics[width=\linewidth]{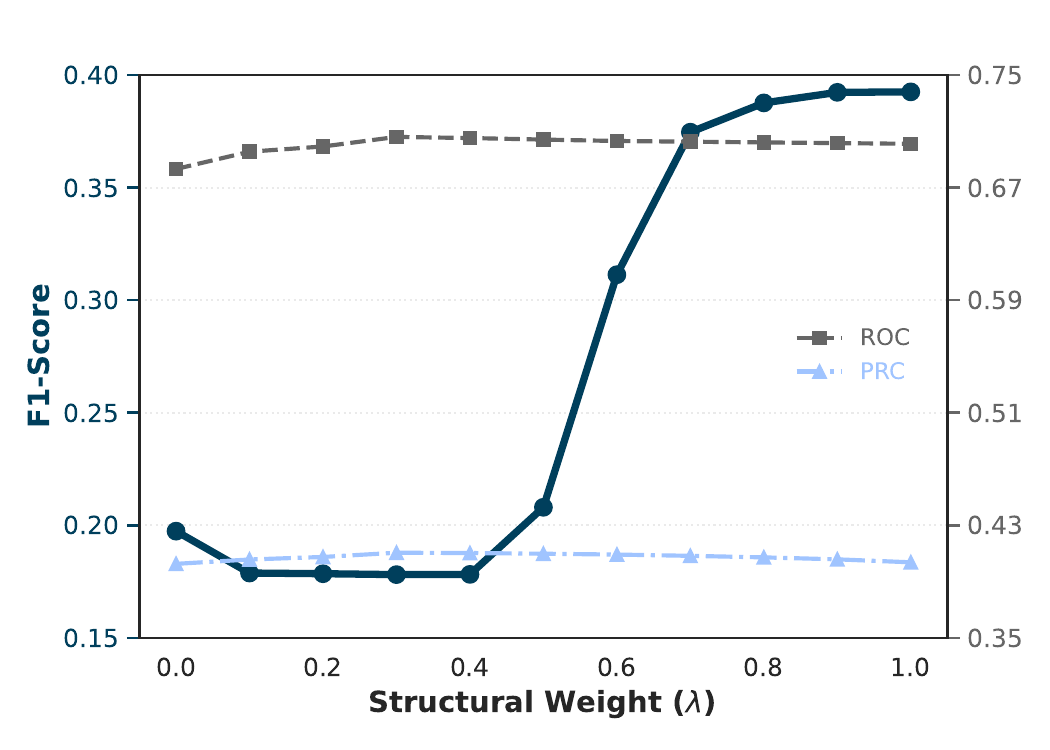}
        \caption{Structural weight $\lambda$.}
        \label{fig:exathlon_sensitivity}
    \end{subfigure}
    \caption{Sensitivity analysis of $\alpha$ and $\lambda$.}
    \label{fig:sensitivity_1x2}
\end{figure}

Experiments on most benchmarks underscore the necessity of introducing the alignment mechanism. We observe a similar trend across all datasets that this alignment constraint enables CAAD to effectively decouple stochastic noise and capture scale-invariant causal logic. By modeling multi-scale latent structural consistencies, it achieves precise forecasting based on history windows, yielding informative residuals that facilitate robust anomaly detection.
Using the PSM dataset as an example, we observe a non-monotonic performance trend characterized by three distinct phases in Fig. \ref{fig:psm_lu_sensitivity}:

\begin{enumerate}[nosep, leftmargin=*]
    \item \textbf{Interference Zone:} At a low weight (e.g., $\alpha=1$), the alignment constraint is insufficient to enforce a coherent cross-scale relationship. Instead, it introduces gradient interference and leads to a temporary drop in the F1-score. In particular, the AUPRC remains relatively stable, suggesting that even weak $\alpha$ maintains the ability of CAAD to rank anomalous.
    
    \item \textbf{Synergy Zone:} Once $\alpha$ reaches an effective threshold (e.g., $\alpha=3$), structural alignment synergizes with the forecasting task, achieving peak performance in both F1 and AUPRC. Across all benchmarks, the CAAD demonstrates steady performance within this zone, exhibiting its robustness.
    
    \item \textbf{Over-regularization Zone:} When $\alpha$ exceeds a certain threshold and has a large weight (e.g., $\alpha=30$), it suppresses informative local signals and leads to a slight performance decay.

\end{enumerate}


\subsubsection{Effect of Relational Causal Quantification}

To analyze the effectiveness of the relational causal score (RCS), we evaluated the CAAD on all datasets through three lenses: parameter sensitivity, causal pattern shifts, and a qualitative case study.

\textbf{Parameter Sensitivity.}
Besides experiments in Table \ref{tab:lambda_sensitivity}, we also validated the effect of $\lambda$ on Exathlon, a cloud-computing benchmark dataset collected under real workloads, including both injected and natural anomalies \cite{jacob2021exathlon}. In Fig. \ref{fig:exathlon_sensitivity}, when the structural weight exceeds 0.5, we observe a sharp increase in the $F_1$-score while ROC and PRC remain stable. The relational causal score enhances detection precision by filtering out pseudo-anomalies, and the optimal precision reaches 0.68, nearly a four-fold increase compared to the dynamic causal score (DCS) alone. Although the DCS is highly sensitive to signal fluctuations, the RCS acts as a structural stabilizer. By verifying that the underlying causal logic remains stable, the RCS effectively filters out false anomalies caused by random numerical oscillations, while maintaining a consistent recall rate.

\begin{table}[t]
\centering
\footnotesize
\setlength{\tabcolsep}{3.5pt}
\begin{tabular}{llcccccc}
\toprule
Dataset & Metric & $\lambda{=}0.0$ & $0.2$ & $0.4$ & $0.6$ & $0.8$ & $1.0$ \\
\midrule
\multirow{2}{*}{SWaT}
 & F1     & 0.9517 & 0.9517 & 0.9543 & 0.9543 & 0.9596 & \textbf{0.9622} \\
 & AUPRC  & 0.9738 & 0.9761 & \textbf{0.9762} & 0.9756 & 0.9740 & 0.9724 \\
\midrule
\multirow{2}{*}{PSM}
 & F1     & 0.3379 & 0.3379 & 0.3379 & 0.3379 & 0.3379 & \textbf{0.3379} \\
 & AUPRC  & \textbf{0.5664} & 0.5663 & 0.5662 & 0.5661 & 0.5659 & 0.5657 \\
\midrule
\multirow{2}{*}{SMD}
 & F1     & 0.1698 & 0.1698 & 0.1698 & 0.1698 & 0.1698 & \textbf{0.1698} \\
 & AUPRC  & 0.0888 & 0.0889 & 0.0890 & 0.0891 & 0.0891 & \textbf{0.0892} \\
\midrule
\multirow{2}{*}{CATSv2}
 & F1     & 0.3300 & 0.3301 & 0.3300 & 0.3302 & 0.3304 & \textbf{0.3304} \\
 & AUPRC  & \textbf{0.3209} & 0.3209 & 0.3209 & 0.3209 & 0.3208 & 0.3208 \\
\bottomrule
\end{tabular}
\caption{Performance trends as the gating strength $\lambda$ increases across all benchmarks, where $\lambda=1.0$ consistently yields optimal or near-optimal F1-scores.}
\label{tab:lambda_sensitivity}
\end{table}

\begin{figure}[t]
    \centering
    \includegraphics[width=0.9\linewidth]{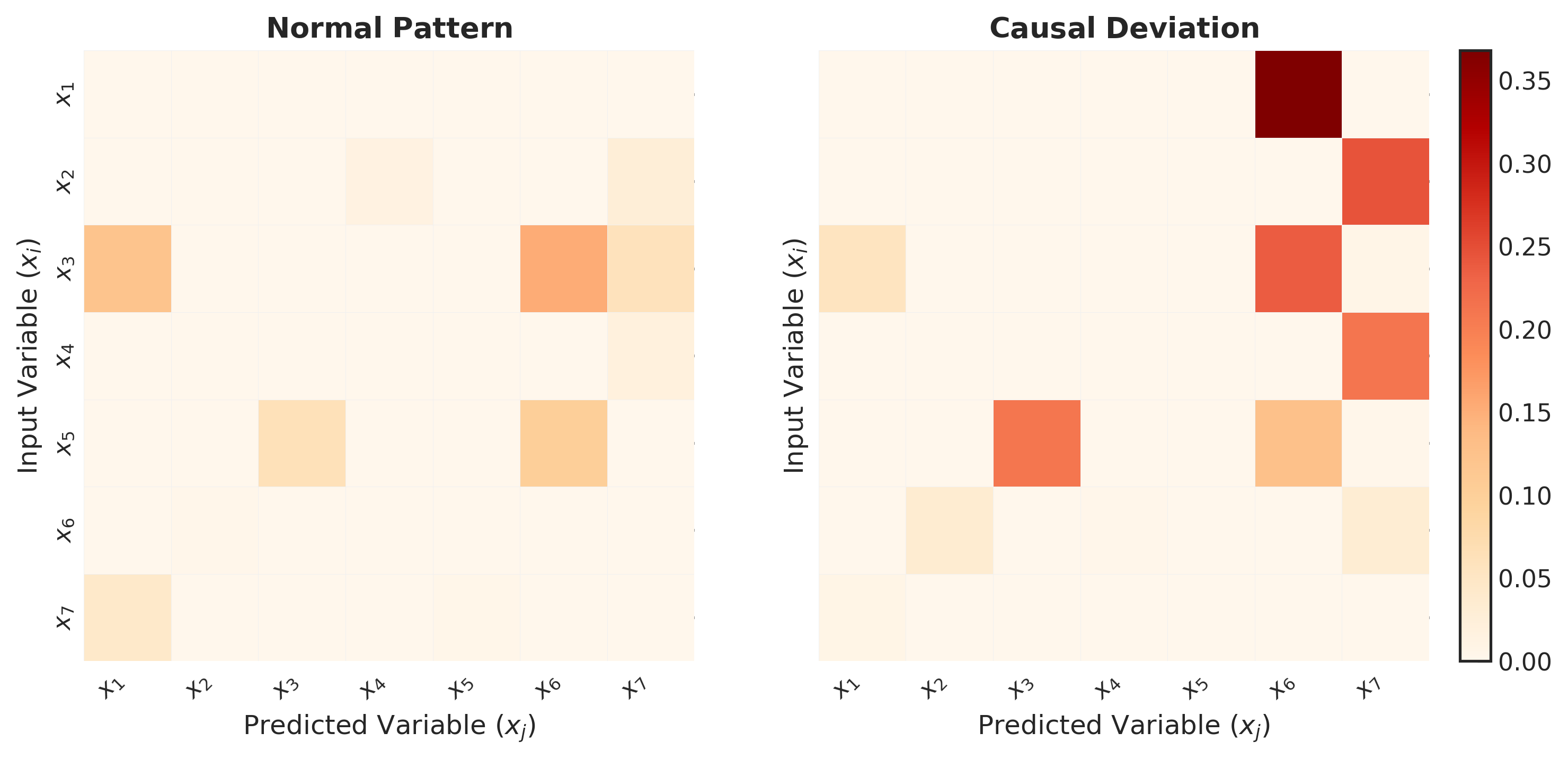}
    \caption{The contrast between normal (left) and anomalous (right) windows validates the RCS's effectiveness in capturing structural logic disruption.}
    \label{fig:structural_2}
\end{figure}

\textbf{Causal Pattern Deviation Analysis.}
Heatmaps in Fig. \ref{fig:structural_2} provide a more intuitive visualization of the effectiveness of RCS. Each cell $(i, j)$ represents the intensity of the causal shift from variable $i$ to $j$ relative to the normal reference matrix $A_{norm}$. In normal windows (left), the heatmap is sparse and pale, indicating a stable topology. However, in anomalous windows (right), deep-colored cells indicate a disruption or alteration in the causal chain. Moving beyond simple numerical outliers, these structural deviations provide interpretable physical evidence that can explain why the system's logic has failed. 

\begin{figure}[t]
    \centering
    \includegraphics[width=0.9\linewidth]{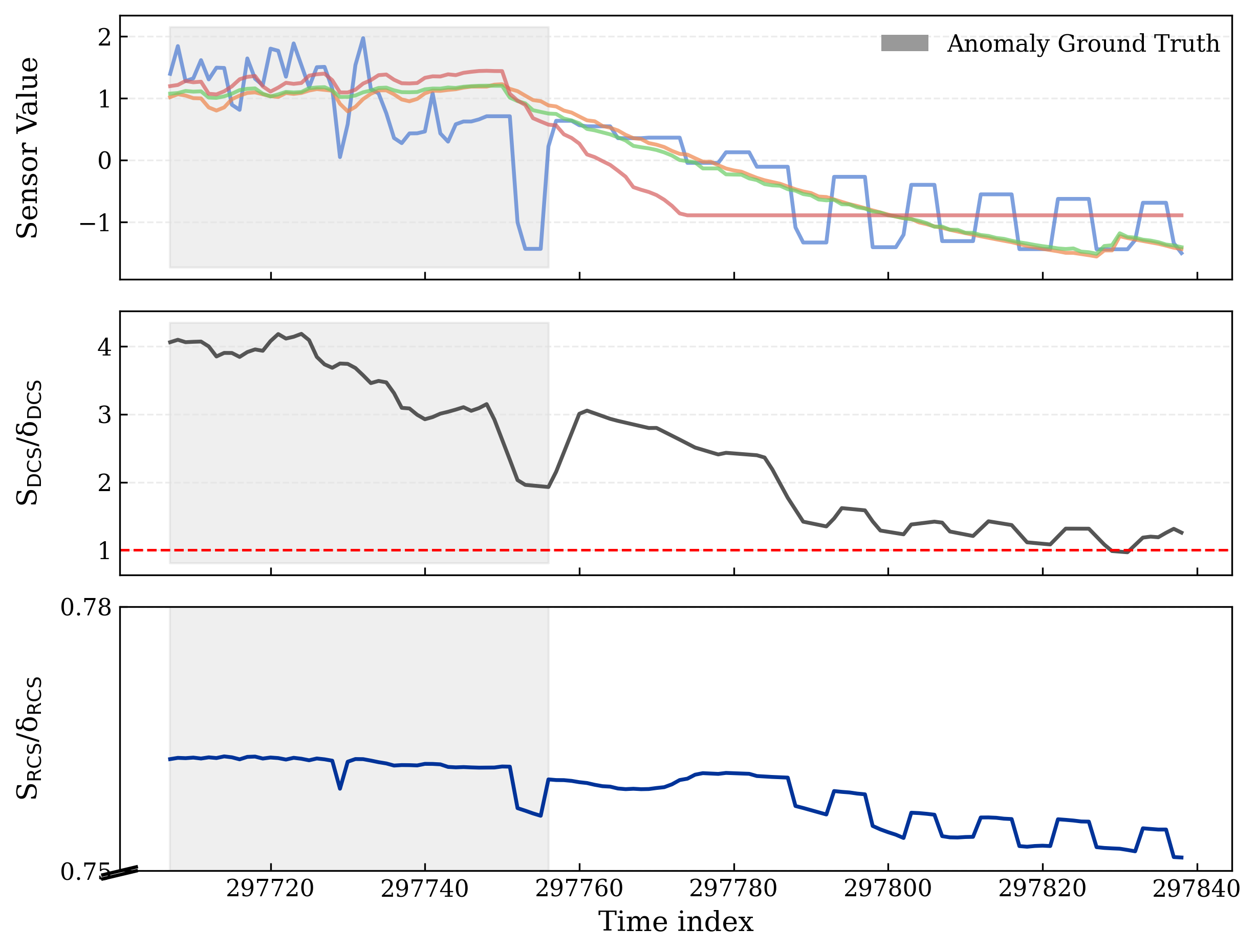}
    \caption{A Case Study on Exathlon, demonstrating RCS's capability of suppressing spurious alerts.}
    \label{fig:structural_3}
\end{figure}

\textbf{Case Study.}
A case study in Fig. \ref{fig:structural_3}  highlights the application scenarios of RCS and the practical benefits of the asymmetric fusion mechanism. The three-tier plot compares the raw signal, DCS, and RCS against their respective validation thresholds.

A common failure mode for prediction-based models is the ``aftershock'' effect: Even after a raw signal returns to its normal range, the prediction residuals (DCS) often remain elevated, triggering persistent false alarms. As shown in Fig. \ref{fig:structural_3}, even after the raw signal leaves the gray anomalous zone and returns to its normal range, the DCS persistently remains above the threshold. In contrast, the RCS stays flat and stable throughout this period because the physical causal logic has not shifted. The weighted fused score suppresses these residual DCS spikes and filters false anomalies, which explains why the structural weight can dramatically boost precision.


\section{Conclusion}
In the present work, we proposed the novel framework \textbf{C}ausality-\textbf{A}ware \textbf{A}nomaly \textbf{D}etection (CAAD), which assumes anomaly detection as a continuous validation of causal dependency. This architecture employs a multi-scale alignment to capture the normal causal patterns across different temporal resolutions, quantifying the causal deviation from dynamic evolution and relational topology perspectives, and incorporating an asymmetric gating mechanism for final anomaly determination. Extensive experiments are conducted on well-recognized public benchmarks and demonstrate that the CAAD outperforms the majority of baselines. For future exploration, we consider expanding the root cause diagnosis function and enhancing the generalization capability across diverse multivariate datasets.

\bibliographystyle{ACM-Reference-Format}
\bibliography{reference}

\appendix
\section*{Appendix}
The appendix is organized as follows:
\begin{itemize}
    \item Appendix A: Implementation Details.
    \item Appendix B:  Additional Experiments.
\end{itemize}

\section{Implementation Details}

\subsection{Multi-scale Windows.}
For PSM, SMD, and CATSv2, the fine-grained window size is set to 60, 100, and 100 respectively, with $k=5$ consecutive fine windows aggregated via mean pooling to construct each coarse-grained sequence. For SWaT, minute-level ($w=60$) and hour-level ($w=24$) resampled sequences are aligned via time-based index mapping. Overlapping sliding windows with stride=1 are used throughout.

\subsection{Model Architecture.}
Both fine and coarse branches use a 2-layer Transformer encoder with $d_{\text{model}}=64$, independently parameterized. The mapping function $M$ is a deterministic index function with no learnable parameters. The top-$K$ parameter in \eqref{eq:dcs}, set to $K=\max(1, \lceil d \times 0.1 \rceil)$, is determined solely by dataset dimensionality without per-dataset tuning. The asymmetric gating threshold $\tau$ in \eqref{eq:gating} is set to $q_{90}$, selected after sensitivity analysis across $\tau \in [q_{70}, q_{99}]$. The aggregation window length $T$ in \eqref{eq:gc} equals the fine-grained window size, aggregating gradients across all timesteps $\tau \in [1, T]$ simultaneously. This naturally captures causal influences at all lags without explicit lag selection, with gradient magnitude reflecting each lag's contribution.

\subsection{Training.}
All models are trained for 50 epochs using AdamW with learning rate $3\times10^{-4}$ and batch size 32.

\subsection{Inference Latency.}
CAAD requires $O(d)$ backward passes per window via Vector-Jacobian Products (VJP), where each pass populates one column of the $d \times d$ causal matrix. Table~\ref{tab:latency} reports measured latency and peak memory across datasets.

\begin{table}[h]
\centering
\caption{Inference latency and peak memory usage.}
\label{tab:latency}
\begin{tabular}{lcc}
\toprule
Dataset & Latency (ms) & Peak Memory (MB) \\
\midrule
CATSv2  & $1.25 \pm 0.05$ & 95.6 \\
PSM     & $2.17 \pm 0.05$ & 64.2 \\
SWaT    & $3.19 \pm 0.04$ & 39.1 \\
SMD     & $4.08 \pm 0.25$ & 97.8 \\
\bottomrule
\end{tabular}
\end{table}
Per-window latency stays under 5ms with peak memory under 100MB, supporting real-time deployment with stride-1 sliding windows and no batch accumulation required.

\section{Additional Experiments}

\subsection{Robustness of $M_n$.}
Under normal operating conditions, causal structure in 
industrial CPS is empirically stable: CAROTS \cite{kim2025causality} reports pairwise cosine similarity above 0.91 across quarterly segments on SWaT, WADI, and PSM, confirming that the causal relationships remain consistent over extended periods. 

To further validate robustness under calibration noise, we validate robustness under two contamination protocols (Gaussian noise injection and anomalous window injection) with contamination ratios from 0\% to 30\% (3 seeds each). As shown in Table~\ref{tab:contamination}, AUROC remains stable at 0.9936 across all conditions, and AUPRC shows only a marginal decrease at 30\% contamination. This robustness is theoretically grounded: $M_n$ is constructed via median aggregation, which has a 50\% breakdown point guarantee.

\begin{table}[h]
\centering
\caption{Performance under $M_n$ contamination on SWaT.}
\label{tab:contamination}
\begin{tabular}{lcc}
\toprule
Contamination Ratio & AUROC & AUPRC \\
\midrule
0\%  & 0.9936 & 0.9314 \\
10\% & 0.9936 & 0.9289 \\
20\% & 0.9936 & 0.9251 \\
30\% & 0.9936 & 0.9219 \\
\bottomrule
\end{tabular}
\end{table}

\subsection{Sensitivity Analysis of $\tau$ and Top-$K$.}
For the asymmetric gating threshold $\tau$ in \eqref{eq:gating}, performance remains highly stable across $\tau \in [q_{70}, q_{99}]$: AUROC fluctuation is within 0.004 on SWaT, with negligible variation on remaining datasets. For top-$K$ in \eqref{eq:dcs}, robustness is consistent across all four datasets, confirming that $K=\max(1, \lceil d \times 0.1 \rceil)$ generalizes without per-dataset tuning.

\subsection{SMD Per-Machine Breakdown.}
SMD comprises 28 machines with substantial heterogeneity in causal coupling strength. Table~\ref{tab:smd} shows the top-3 and bottom-3 machines by AUROC, with a mean AUROC of 0.7812 across all machines.

\begin{table}[h]
\centering
\caption{SMD per-machine AUROC (top-3 and bottom-3).}
\label{tab:smd}
\begin{tabular}{lc}
\toprule
Machine & AUROC \\
\midrule
machine-2-8 & 0.9757 \\
machine-2-3 & 0.9155 \\
machine-2-7 & 0.9087 \\
\midrule
machine-3-7 & 0.6037 \\
machine-3-3 & 0.6274 \\
machine-3-2 & 0.5250 \\
\bottomrule
\end{tabular}
\end{table}

Performance tracks causal coupling strength: machines with denser inter-variable dependencies benefit from causal consistency verification, while sparsely-coupled machines challenge both DCS and RCS. Notably, SMD performance is uniformly low across all methods under TAB's unified protocol \cite{qiu2025tab} (TimesNet F1=0.191, iTransformer F1=0.189, CAAD F1=0.170), indicating that gradual statistical drift in server metrics is a universal challenge rather than a specific limitation of causal approaches. In CAAD, the asymmetric gating in \eqref{eq:gating} implicitly implements a hybrid strategy: RCS is suppressed when structural deviations remain within normal range, naturally reducing to pure DCS in sparse causal settings. 

\subsection{Additional Baseline Comparisons.}
Methods relying on point-adjusted F1 are excluded from direct comparison, as this protocol credits entire anomaly segments if any single point is detected, systematically overestimating performance~\cite{kim2022towards}. We instead compare on threshold-free AUROC on SWaT against methods using compatible evaluation protocols:

\begin{table}[h]
\centering
\caption{SWaT AUROC comparison with related methods.}
\label{tab:baseline}
\begin{tabular}{llc}
\toprule
Method & Type & AUROC \\
\midrule
MTGFlow~\cite{zhou2023mtgflow} & Graph-based & 0.848 \\
CAROTS~\cite{kim2025causality} & Causality-aware & 0.852 \\
GCAD~\cite{liu2025gcad} & Graph + Granger causality & 0.869 \\
\textbf{CAAD (ours)} & Graph + Causality + Multi-scale & \textbf{0.994} \\
\bottomrule
\end{tabular}
\end{table}

Each step adds a component aligned with CAAD's design, with the largest gain attributed to multi-scale temporal alignment: removing it reduces SWaT F1 from 0.952 to 0.784 ($-$17\%), as confirmed by Table \ref{albation}.





\bibliographystyle{ACM-Reference-Format}

\end{document}